\documentclass[10pt,twocolumn,letterpaper]{article}

\usepackage{algorithm}
\usepackage{algpseudocode}
\usepackage{breqn}

\usepackage{booktabs}

\usepackage{hyperref}

\usepackage{iccv}              % To produce the CAMERA-READY version
\definecolor{iccvblue}{rgb}{0.21,0.49,0.74}
\def\paperID{*****} % *** Enter the Paper ID here
\def\confName{ICCV}
\def\confYear{2025}

\title{Dynamic-Aware Video Distillation: Optimizing Temporal Resolution Based on Video Semantics}

\author{
 \textbf{Yinjie Zhao$^{1,2,3}$, Heng Zhao$^{1,2,4}$, Bihan Wen$^{3}$, Yew-Soon Ong$^{1,2,4}$, Joey Tianyi Zhou$^{1,2}$}\\
$^1$CFAR, Agency for Science, Technology and Research (A*STAR), Singapore\\
$^2$IHPC, Agency for Science, Technology and Research (A*STAR), Singapore\\
$^3$School of EEE, Nanyang Technological University, Singapore\\
$^4$CCDS, Nanyang Technological University, Singapore\\
}

\begin{document}
\maketitle
\begin{abstract}

With the rapid development of vision tasks and the scaling on datasets and models, redundancy reduction in vision datasets has become a key area of research. To address this issue, dataset distillation (DD) has emerged as a promising approach to generating highly compact synthetic datasets with significantly less redundancy while preserving essential information. However, while DD has been extensively studied for image datasets, DD on video datasets remains underexplored. Video datasets present unique challenges due to the presence of temporal information and varying levels of redundancy across different classes. Existing DD approaches assume a uniform level of temporal redundancy across all different video semantics, which limits their effectiveness on video datasets. In this work, we propose Dynamic-Aware Video Distillation (DAViD), a Reinforcement Learning (RL) approach to predict the optimal Temporal Resolution of the synthetic videos. A teacher-in-the-loop reward function is proposed to update the RL agent policy. To the best of our knowledge, this is the first study to introduce adaptive temporal resolution based on video semantics in video dataset distillation. Our approach significantly outperforms existing DD methods, demonstrating substantial improvements in performance. This work paves the way for future research on more efficient and semantic-adaptive video dataset distillation research.

\end{abstract}    
\section{Introduction}
\label{sec:intro}

\begin{figure}[ht]
    \centering
    % Specify width as a percentage of the text width, or use height if preferred
    \includegraphics[width=1\columnwidth, trim=55 50 50 30]{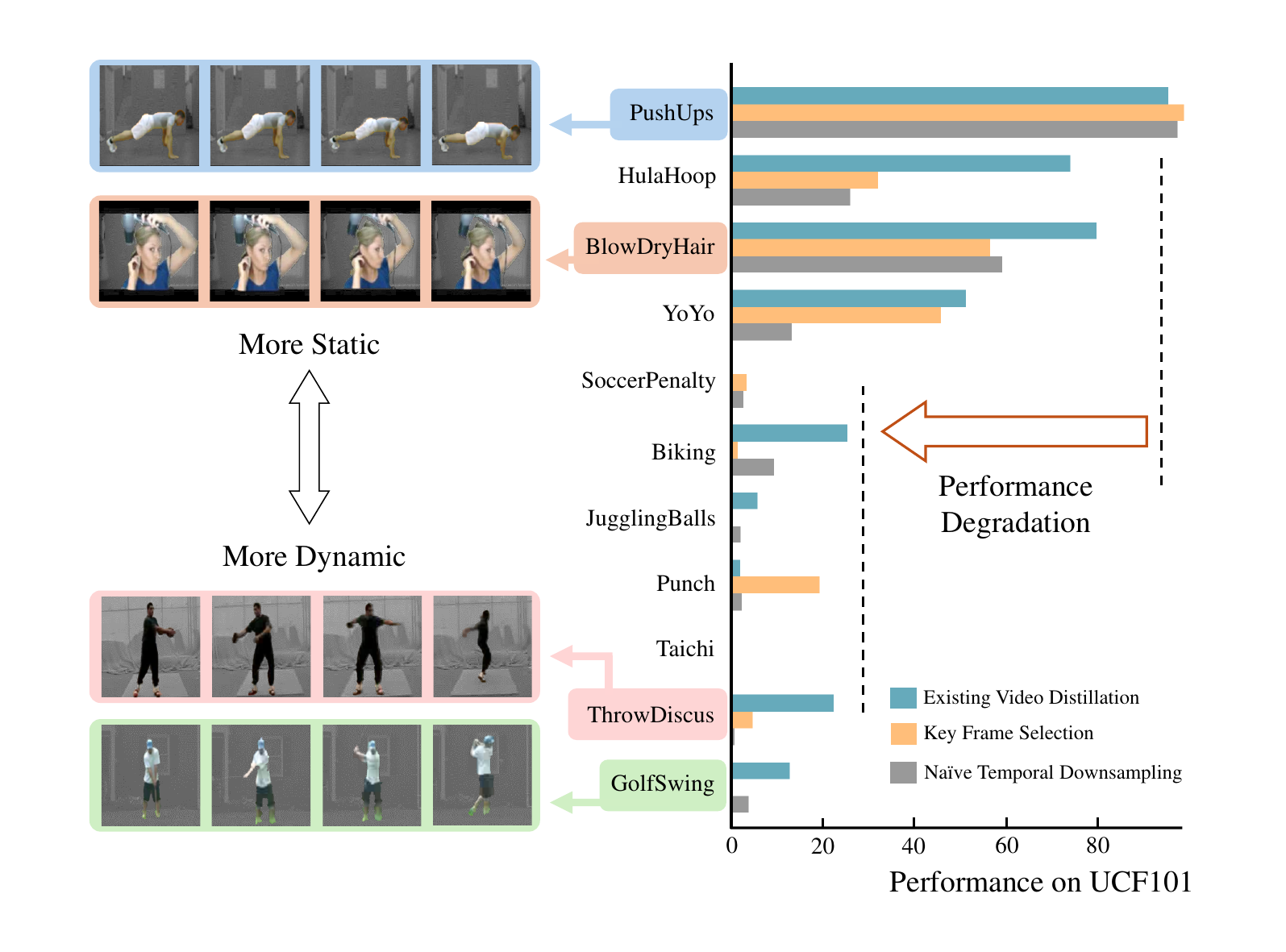}
    \caption{Performance degradation on more dynamic video classes. When constructing a compact video dataset, conventional temporal downsampling and existing video dataset distillation methods often suffer from significant performance degradation, particularly in more dynamic video classes. This limitation arises because these approaches enforce a fixed number of synthetic frames across all video classes, failing to address various levels of redundancy corresponding to different video semantics. Consequently, the representational capacity of the synthetic dataset is damaged, compromising its overall effectiveness.}

    \label{fig:motive}
\end{figure}

% 1. Importance of DD and Video DD

% How redundant are the datasets, and why it is important to reduce redundancies in the dataset
As the development of computer vision research, large-scale datasets \cite{dst4,dst5,dst6,dst7} and vision foundation models \cite{vid_found_mod1,vid_found_mod2, vid_found_mod3}, have been emerging rapidly, meanwhile consuming a significantly higher level of computational resources. Despite higher performance has been continuously achieved, it is also becoming more difficult for those achievements to be replicated and deployed in a computationally light-weighted setting. Such circumstances largely limit the practical value of research results in the computer vision community. As a result, efficiency in data storage and training has become a key challenge and is attracting a rising amount of attention from the research community. 

% DD aims to reduce redundancy and Video DD is important
Dataset Distillation (DD) \cite{dd1} is a task that jointly improves efficiencies in data storage and training process. DD aims to generate a compact synthetic dataset with significantly smaller size compared to the original dataset, while maintaining a comparable training performance. Therefore, DD process could be regarded as a process of redundancy reduction of the original dataset. Initially proposed and explored on image-level datasets, DD has achieved significant improvement, continuously approaching the performance of the original datasets. However, DD on other data modalities remains under-explored. 

Video Understanding is a natural extension from static images into higher dimension with extra temporal information \cite{redun9}, and the spatial and temporal semantics of videos could require separate encoding pathways \cite{redun8}. However, such characteristics of video also comes with much higher level of redundancies. When existing DD approaches are directly applied onto video datasets, large performance degradations are observed \cite{dd12}. This indicates that redundancy reduction in video datasets need to be addressed in novel ways that is different from DD on image-level datasets.

% 2. Nature of Video's diverse level of Temporal Redundancies

% Videos are more challenging due to temporal dimension

Compared to image-level datasets, video datasets are largely affected by its unique challenges along the extra time dimension, especially by the problem of temporal redundancy. Temporal redundancy could be understood as the percentage of frames that could be ignored without compromising the semantic meaning of the video. Since video dataset distillation aims to learn a compact representation of the original dataset, reducing temporal redundancy in videos is a highly necessary and crucial problem. 

% Temporal dimension characteristics are highly related to semantics, including redundancies
The key challenge of temporal redundancy reduction is: videos with different semantic meanings have different level of temporal redundancies. On one hand, if the semantic meaning of the video corresponds to slower events, a high percentage of frames could be ignored, e.g. \textit{Push ups} or \textit{Blowing Dry Hair}. For these slow events, even one single frame could convey the semantic meaning of them, and temporal information rarely matters. This means it only requires a very low temporal resolution. On the other hand, when it comes to faster events such as \textit{Golf Swing} or \textit{Throwing Discus}, it is extremely difficult to determine the percentage of frames that could be ignored. This is because those events corresponds to significantly more dynamic semantic meanings, and ignoring a small part of the video could cause the semantic meaning to be largely compromised. In this case, any representation of the video must use a higher temporal resolution. Therefore, to obtain a highly compact representation of video datasets, it would be necessary to adaptively apply different temporal resolutions to videos of different semantic meanings.

% It is important to adaptively reduce redundancies given different semantic meanings, otherwise, it will cause obstacle to extract useful information in videos 

% 3. Existing approaches cannot solve the problem above 

% Exisitng approaches sucks in dynamic contents
% They sucks because they apply the same temporal resolution to all classes of videos

As shown in Fig.\ref{fig:motive}, we noticed that the performance of existing redundancy reduction approaches, including the state-of-the-art DD \cite{dd12} method on video datasets, is mainly contributed by more static classes videos. We observed severe performance degradation when they are applied to videos with more dynamic semantic meanings. This is because existing approaches naively reduce all videos to the same temporal size, ignoring different levels of temporal redundancies of the original videos.Therefore, to address such severe performance degradation, it is necessary to optimize over a learnable temporal resolution of the synthetic video adaptively across videos of different semantic meanings. 

% 4. Our work. If existing approaches cannot solve it, how do we solve it?

% Our contribution
In this work, we are the first to study the problem of a learnable temporal resolution for different semantical classes of synthetic videos. Due to a large searching space for temporal resolution across the classes, we propose to utilize Reinforcement Learning to address the problem of searching for the optimal temporal resolution. Our contributions can be summarized by the following descriptions:

\begin{itemize}
\item We are the first to propose and investigate the problem of different levels of temporal redundancies across different video semantical classes in video dataset distillation, a factor largely overlooked by existing dataset distillation approaches.
\item We propose a Reinforcement-Learning mechanism with a teacher-in-the-loop reward function to predict the optimal temporal resolution of the Dataset Distillation process.
\item Our approach achieves state-of-the-art results on video Dataset Distillation task, demonstrating its effectiveness in condensing video datasets with and the importance of adaptive temporal resolutions of video DD.

\end{itemize}
\section{Related Work}

\subsection{Dataset Distillation}

% image-level DD does not work 
Dataset Distillation (DD) \cite{dd1} was initially proposed to remove redundancies from image-level datasets. Most of existing DD approaches focuses on image datasets, where a Instance Per Class (IPC) is given as the storage budget constrain. Approaches such as \cite{dd3, dd2, dd11} consider an optimization perspective and enforce the DD optimization process by aligning the training dynamics between the synthetic dataset and the real one. Other approaches \cite{dd7, dd10, dd13} supervise the distillation with the features matching between the synthetic and the real dataset. Generative DD approaches \cite{dd15, dd16, dd17, dd18,dd21} utilizes the prior knowledge in image generative models to improve the efficiency and effectiveness of the synthesization process. However, the performance of these approaches degrades dramatically when applied onto the video datasets.

% scaling dd to large dataset does not work for video set
Meanwhile, the tremendous computational consumption of video DD is another major challenge. TESLA \cite{dd4} propose an efficient memory mechanism on large-scale image datasets. Approaches like SRe$^{2}$L \cite{dd4_1} and RDED \cite{dd23} guide the redundancy reduction process with a teacher model. Despite their performance on image datasets, the problem of temporal redundancy reduction remains unaddressed. The generalizability of image-level solutions is questionable given the extra temporal dimension of video datasets. VDSD \cite{dd12} naively reduce the information of all videos into a static image and interpolate it to compensate for dynamic information. As a result, VDSD has only a marginal improvement compared to existing image-level DD approaches. This indicates that the unique challenge of temporal redundancy are not yet well addressed.

\subsection{Video Temporal Redundancy}

Redundancy reduction has been a challenge for video related tasks, especially along the temporal dimension. There are works exploring conventional approaches including Video Compression \cite{vid_compress1, vid_compress2} and Key Frame Selection  \cite{key_frame_select3,key_frame_select1, key_frame_select2} from a machine learning perspective. Such approaches reduces memory consumption of videos while maintaining the fidelity of the video pixels. However, these approaches struggles when efficiency improvement is desired for videos as a training dataset for machine learning models.

In Video Understanding and Recognition domain, redundancy reduction is a key factor for performance improvement. Learning to remove less informative content in the video is one solution. \cite{redun1} and \cite{redun4} learns to remove uninformative or static content of the videos. \cite{redun10} aims to remove less representative samples in the dataset.  Another solution could be combining the visual tokens into a more compact form.  \cite{redun6} utilizes a clustering mechanism, while \cite{redun7} proposes dynamic visual tokens to represent multiple vanilla visual tokens in the spatial-temporal dimension. Memory mechanism are explored in latent embedding space. \cite{redun2} applies hierarchical short-term to long-term memory mechanism to maintain a efficient representation of the video, and \cite{redun3} utilized a video Q-former which condense the feature of the video token embeddings into a fixed number of query tokens. These existing works on video tasks have demonstrated the importance of dynamic-awareness and adaptability of temporal redundancy reduction for video related tasks.

\subsection{Reinforcement Learning}

Reinforcement Learning (RL) \cite{RL1,RL2} has been a key domain in machine learning. It has been explored to address problems that requires sequential decision makings and non-differentiable choices \cite{RL9}. Among the RL algorithms, Q-learning algorithm \cite{RL4} is a model-free approach which does not require the modeling of the environment, and has a good mathematical guarantee on optimization convergence \cite{RL3}. Such characteristics of RL algorithms provides good potential on solving temporal redundancy problems of videos. RL algorithms has been applied to remove redundancy from video and enforce the focus on informative contents in the videos. RL frameworks on the tasks of key frame selection \cite{RL5}, video grounding \cite{RL6}, segmentation \cite{RL7} and recognition \cite{RL8} have been explored.
\section{Methodology}

\begin{figure*}[ht]
    \centering
    % Specify width as a percentage of the text width, or use height if preferred
    \includegraphics[width=1\textwidth, trim=0 30 0 30]{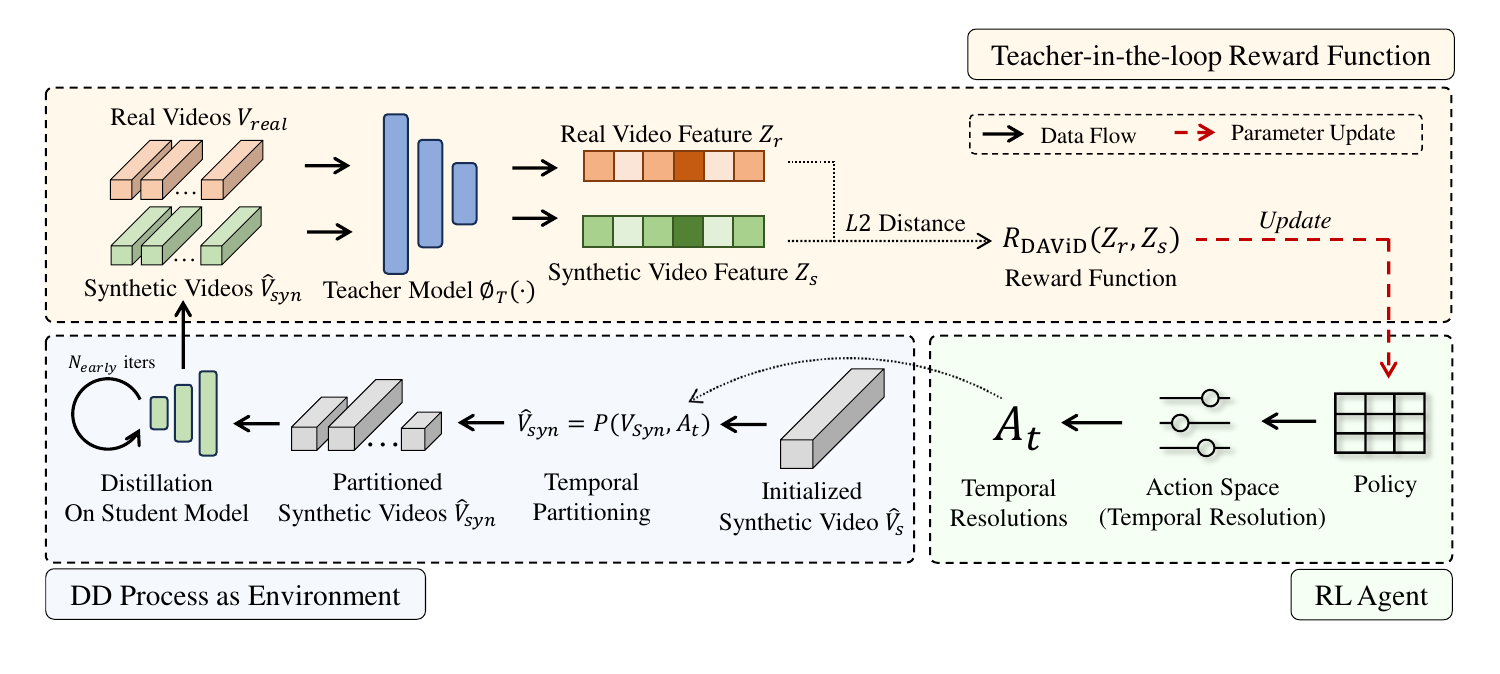}
    \caption{As illustrated in the figure above, the distillation process of DAViD utilizes a reinforcement learning (RL) approach to predict the optimal Temporal Resolution for synthetic data. The learning process of the RL agent is guided by a Teacher-in-the-loop Reward Function, which supervises its policy update. The RL agent predicts the optimal Temporal Resolution for the synthetic videos to adapt to different video semantics. The DD process serves as the environment for the RL agent. We utilize a teacher model to evaluate the feature distance between the synthetic and real data under a given Temporal Resolution, thereby defining the reward function.}

    \label{fig:method}
\end{figure*}

\subsection{Overall Pipeline}

Given the original video dataset $V_{real}$, the task of video DD requires to return a synthetic dataset $V_{syn} \in R^{(M,I,H,W,C)}$, where $M$ is the number of classes and $I$ is Instance Per Class(IPC). Therefore, the target of the task is to condensate the original dataset into the size of $I$ videos per class.

% Over all pipeline is optimizing a hyperparameter of the DD process with RL
Temporal resolution in the video DD optimization process decides the granularity of temporal partitioning of the synthetic video. Therefore, the problem of learning the temporal resolution that maximize the DD performance is an optimization over the hyperparameter of the DD process. 

% Why RL?
% 1. Selecting from discrete space
% 2. Want to minimize search cost (RL gives a good balance between exploration and exploitation). A single search is costly since it takes a DD process.
We propose Reinforcement Learning (RL) as the optimization tool for this problem. Temporal resolutions of the synthetic videos is a finite set of discrete values. On one hand, it is extremely computationally expensive to iterate through all possible choices of temporal resolutions among all classes. On the other hand, the performance of DD process has certain randomness, and might require repeated exploration over the same hyperparameter. Therefore, the optimization tool applied to such task is expected to efficiently estimate the potential gain of different choices in the discrete space with low cost, while maintaining robustness given certain degree of randomness of the gain under a specific choice. Given these circumstances, RL is able to function as an effective optimization tool to address the above challenges. 
   
% DD process as environment, hyperparameter choice as action
As illustrated by Fig.\ref{fig:method}, we propose to formulate the problem of finding an optimal temporal resolution as a Reinforcement Learning problem. Under this formulation, DD process is the environment of the RL agent, returning the gain under different choices of temporal resolutions. The action space is all possible options of temporal resolutions among all classes. The state of such RL process is the temporal resolution applied to the DD process. The updating of the RL agent policy is guided by a Teacher-in-the-Loop reward function as introduced in part \ref{reward_fun}. Detailed optimization process is illustrated in Algo.\ref{alg:DAViD_algo}.

\begin{algorithm}[t]
\caption{\textit{DAViD}($V_{real}$) $\to V_{syn}, A_T$}
\label{alg:DAViD_algo}
\begin{algorithmic}[1]
\State \textbf{Input:} Training Set: $V_{real}$
\State \textbf{Output:} Synthetic Video Set: $V_{syn}$ 
\State \textbf{Utils:} Dataset Distillation Loss: $\mathcal{L}_{DD}$, Teacher-in-the-loop Reward Function: $\mathcal{R}_{TIL}$, Teacher Model: $\Phi_{T}$, Temporal Partitioning Function: $\mathcal{P}$
\State \textbf{Initialize:} Total iterations of RL: $T$, early-stage iterations of DD: $N_{early}, $total iterations of DD: $N$, Q-table: $Q$, Synthetic Video Set: $V_{syn}$, Student Model: $\Phi_{S}$
\\

\Function{Temporal Policy Learning}{}
    \For{$t$ in $T$}
        \State $A_t$ = \textit{take\_action}($Q,\mathcal{A}$)
            \For{$n$ in $N_{early}$}
                \State $\hat{V}_{syn} = \mathcal{P}(V_{syn}, A_t)$
                \State Calculate $\mathcal{L}_{DD}(\hat{V}_{syn},V_{real})$
                \State \textit{DD\_update}($V_{syn}$)
            \EndFor
        \State $\mathcal{R}_{t+1} = \mathcal{R}_{\text{DAViD}}(\Phi_{T}, V_{syn}, V_{real})$
        \State \textit{RL\_update($Q(S_t, A_t)$)}
        
    \EndFor
    \State \textbf{Return:} $Q$
\EndFunction

\\

\Function{Synthesization}{}

    \State $A_{T} = \underset{a}{\mathrm{argmax}}Q(S_{T},a)$
    \For{$n$ in $N$}
        \State $\hat{V}_{syn} = \mathcal{P}(V_{syn}, A_T)$
        \State Calculate $\mathcal{L}_{DD}(\hat{V}_{syn},V_{real})$
        \State \textit{DD\_update}($V_{syn}$)
    \EndFor
    \State \textbf{Return:} $V_{syn}$, $A_{T}$
\EndFunction

\end{algorithmic}
\end{algorithm}

\subsection{Teacher-in-the-Loop Reward Function}
\label{reward_fun}

% We need an reward function refecting the quality of DD optimization under certain Temp Res. 
% Teacher model is able capture meaningful semantics, therefore we use teacher model as an encoder.
% We compare the distance between the synthetic data and the real data

For any RL algorithm, the formulation of the reward function should be considered first. The reward function is expected to drive the agent towards our desired goal. The key challenge of formulating the reward for this RL agent is the lack of ground truth to measure the effectiveness of a DD process. This is because the test set is not visible to the whole framework. For RL frameworks, the lack of explicit gain for actions could cause great challenge in formulating the reward function.

Assuming a well-trained teacher has knowledge on different semantic meanings, it should be able to tell whether features of the synthetic video is meaningful enough for different video semantics. Therefore given such circumstances, we propose to apply a well-trained teacher model as an encoder. The encoder therefore extracts the feature of the synthetic and the real dataset of the same class, which is used to formulate the reward based on the distance between the synthetic and real feature. The formal formulation of the reward function can be expressed as: 
\begin{equation}
\resizebox{0.9\columnwidth}{!}{$\mathcal{R}_{\text{DAViD}}(\Phi_{T}, V_{syn}, V_{real}) = \frac{1}{1+||\Phi_{T}(V_{syn})-\Phi_{T}(V_{real})||_{2}}$}
\end{equation}
Here $\Phi_T$ is the teacher model trained on the original full dataset.

\subsection{Q-learning Guided Dataset Distillation}

% why Q-learning?
% 1. Model-free learning
% 2. Easy to converge

We propose Q-learning as the backbone RL framework for DAViD. On one hand, it is challenging to model the DD process as an environment. This is because of the randomness of the outcome of the DD process, and the computational cost of a DD optimization to converge. On the other hand, we expect the RL algorithm to be easy to converge to the optimal policy.  As a model-free learning process with good theoretical guarantee\cite{RL3}, Q-learning is a suitable RL algorithm for this task.

% Formulation of the RL process. Overall intro of Q-learning
Q-learning process estimates the expected utility (the Q-values) of specific actions under a certain state, and therefore takes actions according to the expected utility. The action space $\mathcal{A}$ is defined as $\mathcal{A} = \{a_1,a_2,...,a_n\}$, where $n$ is the total number of potential options of temporal resolution. Due to the nature of this task, the state space of this RL algorithm is identical to the action space.

% Q-table update
With the reward function defined in \ref{reward_fun}, the update of Q values could be described as the following process: 
\begin{dmath}
$$Q(S_t, A_t) = Q(S_t, A_t) + \alpha * [R_{t+1}+\gamma * max_{a} Q(S_{t+1},a)-Q(S_t, A_t)]$$
\end{dmath}
Here $R_{t+1}=\mathcal{R}_{\text{DAViD}}(\Phi_{T}, V_{syn}, V_{real})$, $\gamma$ is the discount factor for the future rewards, and $\alpha$ is the learning rate of the Q values.

% Taking actions, and Exploitation - Exploration Trade-off

The RL agent take actions with an exploitation-exploration trade-off based on a predefined probability $p$. The process could be formulated as:  

\begin{equation}
\resizebox{0.9\columnwidth}{!}{$
A_t =\textit{take\_action}(Q,\mathcal{A}) =
\begin{cases}
    \underset{a}{\mathrm{argmax}}Q(S_{T},a), & \text{if } x < p \\
    {\mathrm{rand\_choice}} (\mathcal{A}), & \text{if } x \ge p \\
\end{cases}
$}
\end{equation}
Here $x$ is a random variable of a uniform distribution in the interval of $[0,1]$, and $\mathrm{rand\_choice}(\cdot)$ is a random choice process with equal probabilities for each option. Via this mechanism, the RL agent is able to exploit the performance of the estimated best temporal resolution while maintains a $(1-p)$ probability for each action taken to explore new temporal resolution choices.

% Describe Temporal Partitioning
After predicting the temporal resolution $A_t$, we initialize the synthetic videos with noise, and a Temporal Partitioning process is conducted on them. The detailed definition of this process could be expressed as:

\begin{equation}
\resizebox{0.9\columnwidth}{!}{$
\hat{V}_{Syn} = P(V_{Syn}, A_t)
= Temporal\_Resize(Crop(V_{Syn}, A_t))
$}
\end{equation}

Here $Crop(V_{Syn}, A_t)$ is a temporal cropping process. It receives a video and temporal resolution as input, and segments the input video uniformly based on the temporal resolution given. $Temporal\_Resize(\cdot)$ is a simple interleave-repeating process to resize the temporal length of the segments back to the input length of the student model.

\subsection{Efficient Temporal Policy Learning}

Despite the fact that RL algorithm could largely reduce the searching computational workload in the temporal resolution space, it is still challenging for the framework to conduct the distillation process $T$ times on the student model. On one hand, the computational workload of distillation process could be heavy due to the parameter updates of student model. On the other hand, the converging iterations $N$ to update the synthetic videos could be significantly more than the total iteration $T$ of RL pipeline itself, making the searching of optimal temporal resolution extremely inefficient.

Therefore, we set the dataset distillation loss $\mathcal{L}_{DD}$ to be a Distribution Matching (DM) loss\cite{dd10}, and input the synthetic video to be evaluated by the reward function after reaching an early-stage iterations $N_{early}$ of the DD process.

\begin{equation}
    N_{early} = \text{int}(\beta *N), \beta \in (0,1]
\end{equation} 

Here $\beta$ is an hyper-parameter and int$(\cdot)$ is rounding to an nearest integer. Since DM loss does not require an update of the student model, such design could largely improve the efficiency of searching, while still being able to generate meaningful reward signal to the RL agent. 

After the Q-values of the RL agent converges, we update the synthetic video based on the optimal temporal resolution until it converges at total $N$ iterations. This efficient Temporal Resolution Searching process is illustrated in Algo.\ref{alg:DAViD_algo}.

\section{Experiments}

\begin{table*}[t]
\centering
\begin{tabular}{c|c c|c c|c c}
\toprule
 \textbf{Dataset} & \multicolumn{2}{c|}{\textbf{TR-UCF}} & \multicolumn{2}{c|}{\textbf{HMDB51}} & \multicolumn{2}{c}{\textbf{UCF101}} \\ 
\textbf{IPC} & 1 & 5 & 1 & 5 & 1 & 5 \\ \midrule
\textbf{Full DS} &\multicolumn{2}{c|}{45.97 $\pm$ 0.5}& \multicolumn{2}{c|}{28.59 $\pm$ 0.69} & \multicolumn{2}{c}{25.83 $\pm$ 0.0}
\\ \midrule
Random & 6.38 $\pm$ 0.5 & 18.3 $\pm$ 0.1 & 4.6 $\pm$ 0.5 \cite{dd12} & 6.6 $\pm$ 0.7 \cite{dd12} & 3.5 $\pm$ 0.1 & 5.1 $\pm$ 0.5 \\ 
KeyFrame & 6.34 $\pm$ 0.2 & 19.0 $\pm$ 0.4  &  5.1 $\pm$ 0.5 & 7.1 $\pm$ 0.3 & 4.1 $\pm$ 0.2 & 6.1 $\pm$ 0.3 \\ 
DM & 12.7 $\pm$ 0.4 & 20.1 $\pm$ 0.1 & 6.1 $\pm$ 0.2 \cite{dd12} & 8.0 $\pm$ 0.2 \cite{dd12} & 7.5 $\pm$ 0.4 & 13.7 $\pm$ 0.1\\
MTT  & 17.1$\pm$ 0.2 & 27.9 $\pm$ 0.1 & 6.6 $\pm$ 0.5 \cite{dd12} & 8.4 $\pm$ 0.6 \cite{dd12} & 11.4 $\pm$ 0.7 & 18.7 $\pm$ 0.1 \\ 
VDSD & 12.2 $\pm$ 1.0 & 23.6 $\pm$ 0.8 & 8.6 $\pm$ 0.5 \cite{dd12} & 10.3 $\pm$ 0.6 \cite{dd12} & 7.8 $\pm$ 0.0 & 14.6$\pm$0.2\\
DAViD \textbf{(Ours)} & \textbf{26.6} $\pm$ 0.8 & \textbf{38.4} $\pm$ 0.6 & \textbf{12.0} $\pm$ 0.3 & \textbf{20.9} $\pm$ 1.1 & \textbf{15.9} $\pm$ 0.3 & \textbf{23.6} $\pm$ 0.3 \\ \bottomrule
\end{tabular}
\caption{Evaluation on TR-UCF, HMDB51 and UCF101. We propose Temporal-Redundant UCF(TR-UCF) as a new subset of UCF101 dataset, where more dynamic classes with various level of temporal redundancies are collected in this subset. Our approach achieved significant performance advantage on TR-UCF, while obtaining improvement by a large margin on HMDB51 and UCF101 as well. The experimental results indicate the importance of dynamic-awareness during video distillation.}

\label{light_weight_eval}
\end{table*}

\subsection{Datasets}

\subsubsection{Existing Datasets}

To evaluate the effectiveness of our approach, we conduct video DD experiment on the most the classic and commonly used Video Recognition datasets. The evaluation datasets include UCF101\cite{dst1}, HMDB51\cite{dst2}, Something-SomethingV2 (SSv2)\cite{dst3} and Kenetics400 (K400)\cite{dst4}. Existing works\cite{dd12} divide the datasets into light-weight track (UCF101 \& HMDB51) and heavy-weight track (SSv2 \& K400), and we follow this setting for a fair-comparison.

The light-weight track datasets contains relatively less number of classes and real samples in the original dataset. UCF101 dataset focuses on human actions, covering semantic meanings from sports to daily-life activities. With 101 different classes, it has a total of 13,320 videos, and has on average 131 videos per class and average time duration from 2-14 seconds per class. Meanwhile, HMDB51 is the smallest dataset among these, with 6,849 videos in total and 51 classes. The semantic meanings of HMDB51 covers a different set of classes compared to UCF101.

The heavy-weight track datasets have a significantly larger number of classes and total data samples. SSv2 has 220,847 videos in total with 174 classes, mainly focusing ego-centric classes such as human-object interaction. K400 is considered the most challenging dataset viewing from the perspective of efficiency, consisting of 400 classes and in total 306,245 videos samples. The temporal size of the original videos are mostly within 10 seconds.

\subsubsection{Temporal-Redundant UCF (TR-UCF)}
\label{dataset: TR-UCF}

We propose a new split of the UCF101 dataset to study the effect of various temporal redundancy. We categorize the classes based on their level of redundancy, selecting 50 classes with more dynamic semantic meanings, where addressing the temporal redundancy problem is particularly challenging.

To identify videos with more dynamic semantic meanings, we first train the target model on the full dataset. Then, we retrain the model on a static version of the dataset. To produce the static version of the dataset, we randomly select and retain only one frame from each video sample. This modification eliminates the temporal information in the dataset. We then calculate the difference of performance between training on the full dataset and training on the static version. Such calculation could be formulated as:

\begin{equation}
\resizebox{0.9\columnwidth}{!}{$\Delta = eval(train(\Phi_S, V_{real})) - eval(train(\Phi_S, V_{static\_real}))$}
\end{equation}

Here $train(\cdot)$ is the training process that returns the trained model, and $eval(\cdot)$ is the evaluation process that returns the class-wise performance of a given model on the test set. $V_{static\_real}$ is the static version of the dataset as we mentioned. The higher $\Delta$ value is, the more dynamic is the semantic meaning of the class.

For video datasets, we typically expect a significant drop in training performance when temporal information is removed, as the static version lacks the time-dependent features that are crucial for many video recognition tasks. However, in some classes, $\Delta$ is near zero or even has negative values. We identify these classes as those that do not rely heavily on temporal information for recognition. We rank all the classes in the original UCF101 dataset based on the $\Delta$ values, and use the classes with the highest 50 $\Delta$ values to form a new subset of UCF101: Temporal-Redundant UCF (TR-UCF).

Thus, the TR-UCF framework serves as a robust dataset to evaluate the effectiveness of video dataset distillation, particularly in distinguishing the influence of temporal dynamics on model performance.

\subsection{Implementation Details}

% Student model

We apply Distribution Matching\cite{dd10} approach as the image-level distillation approach in DAViD. The student model is a 3-layer C3D\cite{net1} following the setting of VDSD\cite{dd12}. For a fair comparison, we follow the same data loading shape as VDSD to ensure experimental consistency, with a $112*112$ spatial size for the light-weight track and $64*64$ for the heavy-weight track.

% Optimization
We used SGD optimizer with a learning rate of $10$ and a momentum of $0.95$ for the synthetic videos. The optimization process was conducted on three NVIDIA RTX 3090 GPUs for $5*10^3$ iterations. For the training of the student model on the synthetic dataset, we trained the model for $1 \times 10^3$ iterations with a learning rate of 0.01 and applied early stop technique. The early-stage iteration percentage $\beta$ of DD process in the Temporal Policy learning process is set to be $0.02$. The learning rate for Q-learning $\alpha$ is set to be 0.1, and the discount factor for future reward $\gamma$ is set to be 0.5.

\subsection{Evaluation Metrics}

We utilized standard evaluation metrics, commonly employed in Dataset Distillation, to assess the effectiveness of our synthetic dataset. Specifically, for a fixed Instances Per Class (IPC) setting, we evaluated the performance of the synthetic dataset by measuring the accuracy of the student model on the original test set. Temporal partitioning is applied to the evaluation process as we proposed in DAViD.

It is important to highlight that, in the context of Video Set Distillation, the inherently higher memory requirements limit the range of IPC values that can be evaluated. In particular, higher IPC settings such as $IPC=10$ or $IPC=50$ were not tested in our experiments due to memory constraints. We plan to explore these higher IPC settings in future work, particularly through the development of more memory-efficient Video Set Distillation methods.

\subsection{Comparison with Existing Methods}

\subsubsection{Light-weight Track}
% Table 1 Comparison on light weight track
% Descibing the significant improvement
As illustrated in Tab.\ref{light_weight_eval}, our approach has achieved improvement by a large margin. On TR-UCF, our approach achieved a performance more than twice of VDSD\cite{dd12}. With $IPC=5$ in TR-UCF, out approach has achieved 83.5\% performance of the full dataset. On HMDB51, our approach also achieved more than 50\% improvement compared to the previous SOTA. For UCF101 with a more challenging dataset scale, our approach still hold a strong advantage compared to previous SOTA.

% explain why our approach works much better (especially TF-UCF)
We observed our approach preserved a strong performance on TR-UCF, which is a significantly more challenging subset. This is because TR-UCF, as mentioned in part \ref{dataset: TR-UCF}, contains various level of temporal redundancies and requires strong dynamic awareness. Under such great challenge, the experiment results demonstrated the effectiveness of the proposed learnable temporal resolution framework.

\subsubsection{Heavy-weight Track}
% Table 2 Comparison on heavy weight track

% Descibing the improvement

Our approach shows stronger effectiveness on the heavy-weight track datasets, especially under a higher IPC setting. Heavy-weight datasets such as SSv2 or K400 are extremely challenging. SSv2 mainly focuses on videos with ego-centric semantic meanings, while K400 contains significantly more number of classes. Such characteristics of these datasets largely limits the performance of student models. We observed that our approach obtains a higher advantage under a higher IPC setting, this could be because our approach benefit more from the growth of sample diversity as the IPC increases.

It is worth mentioning that our approach uses Distribution Matching as the backbone, while VDSD's performance shown in this table uses MTT as the backbone, which requires significantly more computational resources.

\subsubsection{Class-wise Analysis}
\begin{figure}[t]
    \centering
    % Specify width as a percentage of the text width, or use height if preferred
    \includegraphics[width=1\columnwidth, trim=20 20 20 20]{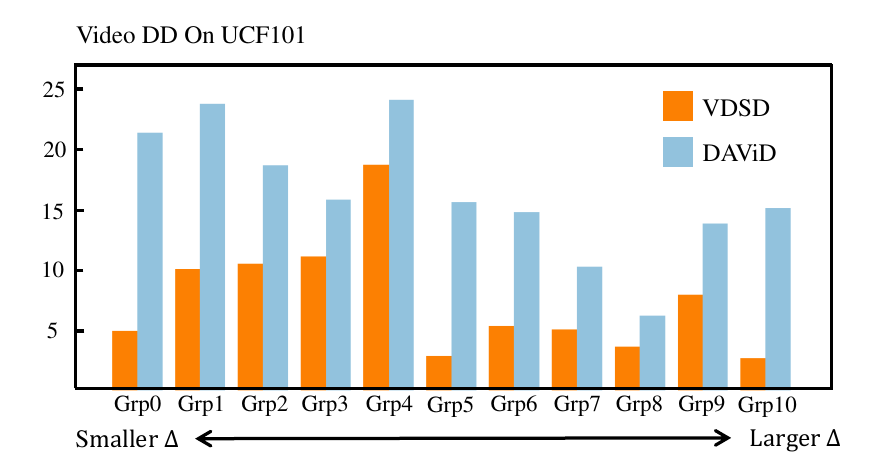}
    \caption{Distillation effectiveness affected by dynamic semantic meanings. As illustrated in this figure, we divided UCF101 dataset into 11 groups according to their $\Delta$ values. From left to right, Group 0 is the most static group of videos and Group 10 is the most dynamic group of videos. Existing approaches only performs relatively well on Group 1-4, and their performance dramatically drops on other groups. In contrast, our proposed method not only dramatically outperforms existing approaches, but also have a balanced performance across all different semantic meanings.}
    \label{fig_class_wise}
\end{figure}

% Figure 3 Class-wise performance boosting especially on different groups
% describe the meaning and setting of figure 3, and the proof of dynamic-awareness
As illustrated in Fig.\ref{fig_class_wise}, we grouped different classes in UCF101 dataset according to their $\Delta$ values. Group 0 has the smallest $\Delta$, meaning the classes require the least temporal information to be recognized, and vice versa for Group 10. The value of each bar is the average performance among the classes in a certain group. The performance degradation of existing approach is significant from Group 5 to Group 10. In contrast, our approach shows a more balanced result among classes, while achieving a universal improvement across different classes. 

% insights on the limitation of current methods
Furthermore, given different level of dynamic semantics, the difficulty of classes varies significantly. We can come to a conclusion that existing approaches not only largely ignored the problem of various temporal redundancy, but also struggles to deal with hard samples in the dataset. This largely limits the effectiveness and robustness of the DD process. This indicates that the inter-class balance problem requires more attention in the task of video DD.

\setlength{\tabcolsep}{3pt} % Default is 6pt

\begin{table}[t]
\centering
\begin{tabular}{c|c c|c c}
\toprule
 \textbf{Dataset} & \multicolumn{2}{c|}{\textbf{SSv2}} & \multicolumn{2}{c}{\textbf{K400}} \\ 
\textbf{IPC} & 1 & 5 & 1 & 5  \\ \midrule
\textbf{Full DS} &\multicolumn{2}{c|}{29.0 $\pm$ 0.6}& \multicolumn{2}{c}{34.6 $\pm$ 0.5}\\ \midrule
Random \cite{dd12} & 3.3 $\pm$ 0.1 & 3.9 $\pm$ 0.1 & 3.0 $\pm$ 0.1 & 5.6 $\pm$ 0.0  \\ 
DM \cite{dd12} & 3.6 $\pm$ 0.0 & 4.1 $\pm$ 0.0 & 6.3 $\pm$ 0.0 & 9.1 $\pm$ 0.9 \\
MTT \cite{dd12}  & 3.9 $\pm$ 0.1 & 6.3 $\pm$ 0.3 & 3.8 $\pm$ 0.2 & 9.1 $\pm$ 0.3  \\ 
VDSD \cite{dd12} & \textbf{5.5} $\pm$ 0.1 & 8.3 $\pm$ 0.2 & 6.3 $\pm$ 0.1 & 11.5 $\pm$ 0.5 \\
DAViD\textbf{(Ours)} & 3.5 $\pm$ 0.5 & \textbf{9.1} $\pm$ 0.3 & \textbf{7.1} $\pm$ 0.1 & \textbf{13.8} $\pm$ 0.1 \\ \bottomrule
\end{tabular}
\caption{Evaluation on heavy-weight track datasets, SSv2 and K400. These larger-scale datasets contains significantly more classes and more diverse semantic meanings. Despite the great challenges, our approach achieved SOTA performance on the datasets.}
\end{table}

\subsection{Ablation Studies}

In this ablation study, we aim to verify the efficiency of our Temporal Policy Learning mechanism, and the necessity to adapt to different video semantics when aiming to reduce temporal redundancies.

\subsubsection{Temporal Policy Learning Efficiency}

As illustrated in Tab.\ref{tab:gpu_time}, we compared with two baseline settings, Grid Search and naive RL Search. Grid Search means to iterate through the whole Temporal Resolution action space $\mathcal{A}$ with a predefined step length. Naive RL search means the RL agent will wait until the DD process to converge before getting the reward for an action. The RL Search with Early Stage DD is the approach proposed in our DAViD.

We tested the Wall-Clock GPU time on a single NVIDIA RTX 3090 GPU to compare the efficiency of different temporal policy learning process. The results shows that our prposed Efficient Temporal Policy Learning approach is 2 orders of magnitude more efficient compared to Grid Search regarding GPU Time.

\begin{table}[t]
    \centering
    \begin{tabular}{lcc}
        \toprule
        \textbf{\begin{tabular}{c}
       Wall-Clock GPU Time \\ (Hours) 
     \end{tabular}
     }  & \textbf{IPC = 1} & \textbf{IPC = 5} \\
        \midrule
        Grid Search & 96 & 155.8 \\
        RL Search & 48.2 & 81.3 \\
        RL Search w. Early Stage DD & \textbf{0.9} & \textbf{1.6} \\
        \bottomrule
    \end{tabular}
    \caption{Comparison of wall-clock GPU Time for different Temporal Policy learning methods on HMDB51. We aim to verify the contribution of RL Search and the early-stage DD mechanism to the overall temporal policy learning efficiency. The approach we proposed is 2 orders of magnitude more efficient compared to Grid Search.}
    \label{tab:gpu_time}
\end{table}

\subsubsection{Video Semantics Adaptiveness}

% Intro to ablation settings
In this part of ablation study, we aim to verify the effectiveness of the RL guided video distillation pipeline. As illustrated in Fig.\ref{tab:ablation1}, we tested out three different cases. Case A is conducting distillation without the temporal policy guidance of RL, and naturally this case does not allow a learnable temporal resolution. Case B is using a universal temporal policy for all video classes. As shown in this ablation experiment, such universal temporal resolution is not able to adapt different semantic meanings of the video, and it caused an sub-optimal performance. The third case is the full implementation of our DAViD approach.

% What does it mean?
% 1. T-partitioning works but it only gives limited improvement
% 2. Learnable T-Res is the main performance contributor

The results of the ablation indicates the importance of adaptable temporal resolution to different video semantics. Such observation verified the problem we observed in Fig\ref{fig:motive}, and it shows the necessity of dynamic-awareness when conducting video dataset redundancy reduction.

\begin{table}[t]
\centering
% \Large
\begin{tabular}{c c c c}
\toprule
 &\multicolumn{1}{c}{
     \begin{tabular}{c}
       RL Guided \\ Partitioning
     \end{tabular}
     } 
     & 
 \multicolumn{1}{c}{
 \begin{tabular}{c}
        Adaptive to  \\
        Semantics
     \end{tabular}
  } 
 
 & TR-UCF \\
 
 \midrule
 
A &  &  & 15.4 $\pm$ 0.2\\

B & \checkmark &  & 20.0 $\pm$0.5\\

DAViD & \checkmark & \checkmark & \textbf{26.6} $\pm$ 0.8\\

\bottomrule
\end{tabular}
\caption{Adaptiveness to video semantics. Case A utilizes neither RL-guided temporal partitioning nor Temporal Policies adaptive to video semantics, this case is equivalent to a naive image-level dataset distillation approach. Case B with only temporal partitioning does not have a dynamic-awareness of video semantics.}
\label{tab:ablation1}
\end{table}

\subsection{Visualizations}

\begin{figure}[t]
    \centering
    % Specify width as a percentage of the text width, or use height if preferred
    \includegraphics[width=1\columnwidth, trim=35 40 20 45]{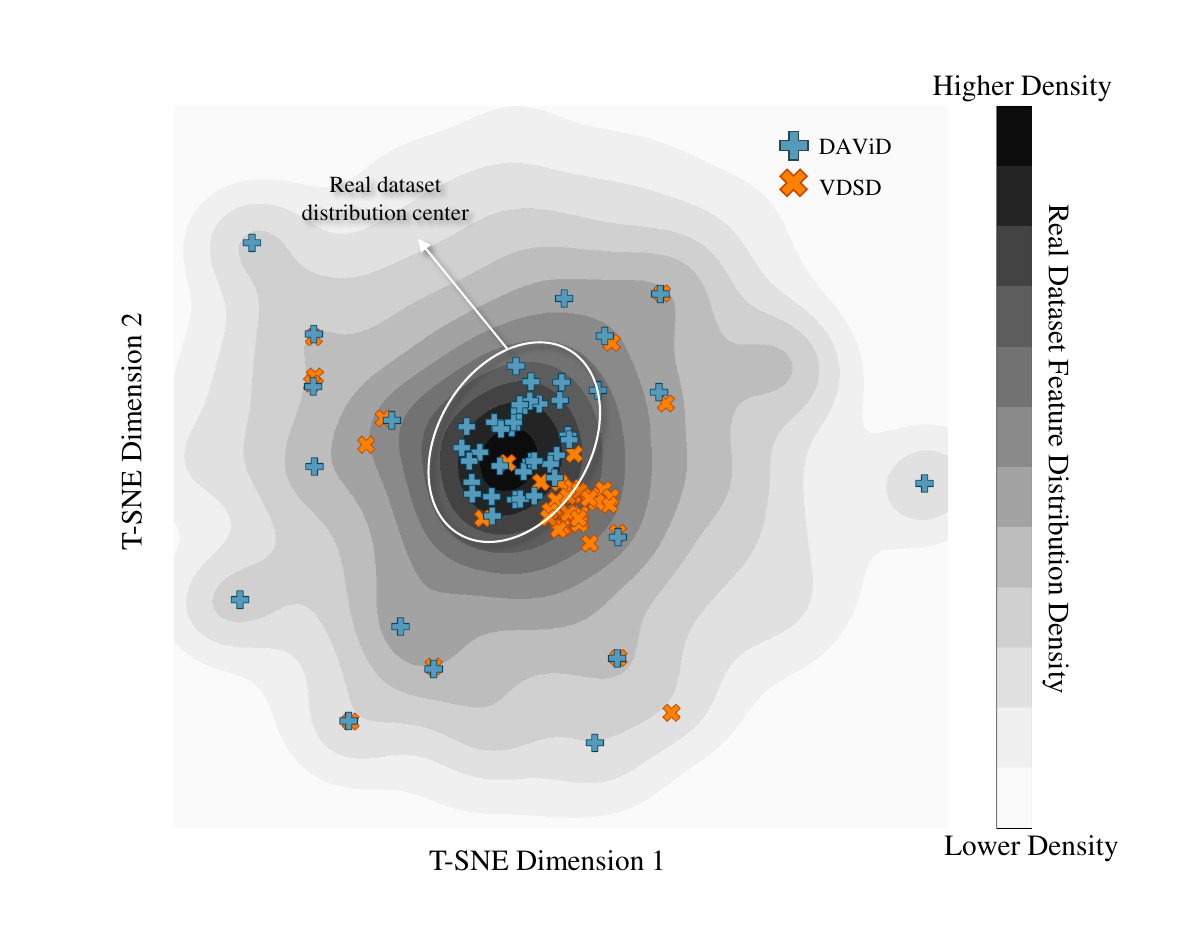}
    \caption{Feature space T-SNE visualization of synthetic videos on TR-UCF dataset. The gray-scale background shows the density of real dataset distributions. Our approach not only has a better representation of majority classes closer to the center of real data feature distribution, but also shows a better diversity representing minority classes on the edge. This indicates that dynamic-awareness is crucial to the representativeness of synthetic video datasets.}
    \label{fig:tsne_fig}
\end{figure}

As illustrated in \ref{fig:tsne_fig}, we visualized the 2D T-SNE similarity of the features among the real dataset videos, the synthetic video produced by VDSD\cite{dd12} and our DAViD. The gray-scale background shows the density of real dataset feature distributions, and each point in the figure is the mean feature of a class of the synthetic data.

On one hand, it is observed that our approach clearly represents the majority classes in the center better, while VDSD produces synthetic videos with a feature distance gap from the real dataset. Such gap in feature distribution could be caused by the loss of temporal information. If most feature preserved by VDSD is the static features, the feature could still have certain similarity to the real dataset, but unable to represent the dynamic feature from the real videos.

On the other hand, for the classes far from the majority, our approach returns a clearly more diverse and representative feature distribution compared to VDSD. The feature visualization clearly shows the importance and effectiveness of preserving temporal information and dynamic-awareness. 

\section{Conclusion}

In conclusion, this is the first work to address the problem of learnable temporal resolution for video distillation tasks. We propose a Dynamic-Aware Video Distillation (DAViD) with a RL framework to adaptively predict the optimal temporal resolution for different video semantics. An efficient temporal policy learning mechanism is proposed. Our approach achieved SOTA performance on multiple video distillation datasets and we achieved two orders of magnitudes higher searching efficiency in temporal resolution space compared to Grid Search. This work paved ways for further studies to more efficient video recognition and redundancy reduction.
{
    \small
    \bibliographystyle{ieeenat_fullname}
    \bibliography{main}
}

\end{document}